\documentclass[12pt]{article}
\usepackage{color}
\usepackage{geometry}
\usepackage{graphicx}
\usepackage{caption}
\usepackage{subcaption}
\usepackage{float}
\usepackage{array}

\usepackage{epsfig}
\usepackage{times}
\usepackage{amsmath}

\usepackage[round]{natbib}
\usepackage{appendix}
\usepackage{amssymb}
\usepackage{multirow}
\usepackage{setspace}
\usepackage{verbatim}
\usepackage{caption}
\usepackage{subcaption}
\usepackage{lineno}

\usepackage[dvipsnames]{xcolor} \definecolor{LightGray}{rgb}{0.97,0.97,0.97}
\usepackage{listings} \lstdefinelanguage{SPARQL}{
  basicstyle=\small\ttfamily,
  backgroundcolor=\color{LightGray},
  columns=fullflexible,
  breaklines=false,
  sensitive=true,
frame=bt,
  aboveskip=1em,
  belowskip=1em,
  xleftmargin=.5em,
  xrightmargin=.5em,
  framexleftmargin=.5em,
  framextopmargin=.5em,
  framexbottommargin=.5em,
  framexrightmargin=.5em,
tabsize = 2,
  showstringspaces=false,
  morecomment=[l][\color{gray}]{\#},       morecomment=[n][\color{blue}]{<http}{>}, morestring=[b][\color{OliveGreen}]{\"},  keywordsprefix=?,
  classoffset=0,
  keywordstyle=\color{Sepia},
  morekeywords={},
classoffset=1,
  keywordstyle=\color{Purple},
  morekeywords={rdf,rdfs,owl,xsd,purl, ev-ont, kwg-ont, geo},
classoffset=2,
  keywordstyle=\color{MidnightBlue},
  morekeywords={
    SELECT,CONSTRUCT,DESCRIBE,ASK,WHERE,FROM,NAMED,PREFIX,BASE,OPTIONAL,
    FILTER,GRAPH,LIMIT,OFFSET,SERVICE,UNION,EXISTS,NOT,BINDINGS,MINUS,a, DISTINCT, VALUES
  }
}

\usepackage{authblk}
\newcommand{\yanlin}[1]{\textcolor{black}{{#1}}}

\definecolor{darkblue}{RGB}{0,0,147}
\usepackage[final]{hyperref}
\hypersetup{
	colorlinks=true, 
	linktocpage=true,
	breaklinks=true, 
	pageanchor=true,
	plainpages=false, 
	hypertexnames=true,
	urlcolor=darkblue, 
	linkcolor=darkblue,
	citecolor=darkblue, 
	pdftitle={EVKG: An Interlinked and Interoperable Electric Vehicle Knowledge Graph for Smart Transportation System},
	pdfauthor={Yanlin Qi, Gengchen Mai, Rui Zhu, Michael Zhang}
	pdfsubject={Semantic Web},
	pdfkeywords={Linked Data, Semantic Web, GIS computation},
	pdfcreator={pdfLaTeX},
	pdfproducer={LaTeX via TeX Live on GNU/LINUX}
}

\usepackage [autostyle, english = american]{csquotes}
\MakeOuterQuote{"}

\usepackage{listings}
\let\oldhat\hat

\renewcommand{\hat}[1]{\oldhat{\mathbf{#1}}}

\title{EVKG: An Interlinked and Interoperable Electric Vehicle Knowledge Graph for Smart Transportation System}

\author[1]{\small Yanlin Qi}
\author[2]{Gengchen Mai}
\author[3]{Rui Zhu}
\author[1]{Michael Zhang}

\affil[1]{Institute of Transportation Studies, University of California, Davis, 95616, California, USA}
\affil[2]{Spatially Explicit Artificial Intelligence Lab, Department of Geography, University of Georgia, Athens, Georgia, 30605, USA}
\affil[3]{School of Geographical Sciences, University of Bristol, Bristol,
	BS8 1SS, United Kingdom}

\date{}
\begin{document}
	
\pagestyle{plain}
\pagenumbering{roman}
\maketitle



\begin{abstract}

\noindent 
\yanlin{Over the past decade, the electric vehicle industry has experienced unprecedented growth and diversification, resulting in a complex ecosystem. To effectively manage this multifaceted field, we present an EV-centric knowledge graph (EVKG) as a comprehensive, cross-domain, extensible, and open geospatial knowledge management system. The EVKG encapsulates essential EV-related knowledge, including EV adoption, electric vehicle supply equipment, and electricity transmission network, to support decision-making related to EV technology development, infrastructure planning, and policy-making by providing timely and accurate information and analysis.
To enrich and contextualize the EVKG, we integrate the developed EV-relevant ontology modules from existing well-known knowledge graphs and ontologies. This integration enables interoperability with other knowledge graphs in the Linked
Data Open Cloud, enhancing the EVKG's value as a knowledge hub for EV decision-making. Using six competency questions, we demonstrate how the EVKG can be used to answer various types of EV-related questions, providing critical insights into the EV ecosystem. Our EVKG provides an efficient and effective approach for managing the complex and diverse EV industry. By consolidating critical EV-related knowledge into a single, easily accessible resource, the EVKG supports decision-makers in making informed choices about EV technology development, infrastructure planning, and policy-making. As a flexible and extensible platform, the EVKG is capable of accommodating a wide range of data sources, enabling it to evolve alongside the rapidly changing EV landscape.}


\end{abstract}
\pagenumbering{arabic}

\section{Introduction}
\noindent

As electric vehicles (EVs) resonate across the global automotive industry, technological innovations and environmental regulatory on gasoline-powered vehicles further accelerate the expansion of EV market \citep{bonges2016addressing}. While global light auto sales lost 8.1\%, the market of battery electric vehicles (BEVs) and plug-in hybrid electric vehicles (PHEVs) increased by 62\% with a total of 4.3 million sales during the first half of 2022 \citep{irle_2022}. Such a trend shows an emphatic global shift to vehicle electrification. The characteristics of the EV market also distinguish itself from the traditional internal combustion engine (ICE) vehicles \citep{jenn2020depth,husain2021electric}.

Firstly, \yanlin{the advent of EVs has brought significant changes to the automobile industry, particularly in terms of vehicle configuration and charger compatibility.  Unlike ICE vehicles, which have a standardized design,   EVs require unique components that are specifically developed for them, such as the battery pack \citep{bonges2016addressing}. This key difference not only affects the way that EVs are designed and manufactured but also has significant implications for the way that they are charged and maintained \citep{das2020electric}.} Such a transition has fostered many established automobile makers to develop new models to support EV, alongside numerous start-ups entering the market. Vehicle configurations, therefore, become diverse across EV manufacturers by a selection of unique makers and models, especially on the battery capacities and charger types. Today, no universal charger or connector type exists in the EV market \citep{center2015electric} and none vehicle is compatible with every type of charger.  For instance, Tesla's Superchargers are exclusively designed to be used on certain Tesla EV models. \yanlin{The diverse configurations of EVs as well as their chargers have created a pressing need for advanced data and knowledge management techniques to organize and interlink this vast amount of information. In this context, the capability of an interlinked data repository such as a knowledge graph to explicitly describe the relationships between different EV models and chargers would prove valuable, as it would provide a solution for managing the complex EV market.}

\yanlin{Secondly, there exists an incompatibility of EV with electric vehicle supply equipments (EVSE).} Unlike filling up a gas tank within a few minutes, replenishing the power of EVs can take a significant amount of time. Most EV charging stations are also decentralized and have a complex structure, operated by diverse providers with different constraints, such as parking limits, membership requirements, charging levels, and connector constraints \citep{rajendran2021comprehensive}. \yanlin{One main challenge posed by such heterogeneity of EV and EVSE systems is the significant variations in their spatial distribution which can cause supply and demand imbalance in certain local regions \citep{carlton2022electric}. EV adoption rates can be influenced by various external factors, such as regulatory mandates, sustainability goals, and user income levels \citep{song2020existing}. These factors often lead to uneven spatial distributions of EV adoption across different spatial scales. However, the installment of EVSEs is struggling to keep pace with the rate of EV adoptions at the current stage \citep{csiszar2019urban}. Additionally, the data format heterogeneity of EV and EVSE makes semantic interoperability impossible. Different types of electric vehicles and charging stations have their own unique features and specifications, which can result in differences in communication protocols and data formats \citep{anil2020electric}. These differences in EV and EVSE characteristics can create barriers to interoperability, making it complicated to charge different types of electric vehicles at different types of charging stations. This data format heterogeneity also makes it complicated for charging infrastructure planning. 
To overcome these challenges, integrating EV and EVSE data into an interlinked data repository such as a knowledge graph is a viable solution to breaking down data silos and gaining a more comprehensive understanding of the overall system. By ensuring data interoperability, it is possible to establish consistent data collection and reporting standards, enabling easier comparisons and insights.}

\yanlin{Meanwhile, it is crucial to carefully select charging station sites to ensure the sustainability of the whole power grid system. Unlike gas stations for refueling ICE vehicles, EVSE infrastructure plays a critical role in the flow of electricity and significantly affects the power grid sustainability \citep{das2020electric}. As the number of EVs increases, the high demand for electricity from EVSEs may overload the electricity system, leading to power blackouts with far-reaching consequences if not managed correctly \citep{dharmakeerthi2012modeling}. In contrast, the power grid may need to control the electricity flow to charging stations to ensure they receive a consistent power supply \citep{yang2014improved}. 
However, in reality, EVSEs management and power grid management are usually operated independently by different agencies without a framework for sufficient data sharing which may lead to unpredictable consequences in the future.
Without integrating data from EVSEs and the power grid, it can be challenging to deploy charging infrastructure strategically. This mismatch between charging infrastructure demand and supply can result in inconvenience for EV users and a slower rate of EV adoption.
Therefore, an integrated data repository (e.g., a knowledge graph) of EVSEs and the power grid data
is also essential for an efficient management of EV charging infrastructure planning, which can accelerate EV adoption and achieve 
sustainable transportation.}

\yanlin{The EV industry is a rapidly evolving and complex field, with numerous stakeholders, technologies, and policies involved \citep{kley2011new}. While there are many open EV datasets available \citep{calearo2021review, amara2021review}, the challenge lies in effectively managing, sharing, and interoperating knowledge across different systems. This complexity makes it challenging to integrate and manage diverse sources of information and knowledge related to the EV industry, which highlights the need for a comprehensive and cohesive EV knowledge management system.
A smart EV knowledge management system is crucial for the efficient integration and analysis of data from various sources, including EV manufacturers, charging station operators, utilities, and government agencies. 
It can provide valuable insights into key industry trends, challenges, and opportunities, and help to achieve data and semantic interoperability , and facilitating communication and collaboration among different stakeholders in the EV industry \citep{cao2021electric}. Moreover, such a system can support decision-making related to EV technology development, infrastructure planning, and policymaking by providing timely and accurate information and analysis.}

\yanlin{To address the challenges in the EV industry, knowledge graphs (KGs) have emerged as a promising solution \citep{noy2019industry}}.  As a novel data paradigm, knowledge graphs are a combination of technologies, terminologies, and data cultures for densely interconnecting (Web-scale) data across domains in a human and machine-readable format \citep{bizer2011linked,janowicz2022know}. With an ontology, or so-called knowledge graph schema, to encode the terminology semantically, KGs also foster interoperability across different domains \citep{hitzler2021review,zhu2022covid}. 
Today, open KGs such as \textit{DBpedia} \citep{auer2007dbpedia}, and \textit{Wikidata} \citep{vrandevcic2012wikidata,vrandevcic2014wikidata} are considered valuable assets for exploiting a broad scope of cross-domain linked data. 
Other than these general-purpose graphs, we also have various large-scale geospatial knowledge graphs such as 
\textit{GeoNames}\footnote{\url{https://www.geonames.org/}} \citep{ahlers2013geonames} and \textit{KnowWhereGraph}\footnote{\url{https://knowwheregraph.org/}} \citep{janowicz2021knowwheregraph,janowicz2022know} which have shown unique superiority in uplifting environmental intelligence as well.

 In this work, inspired by the challenges faced by EV charging systems and recent advancements in KG, we develop an EV-centric knowledge graph, called EVKG, that serves as an interlinked, cross-domain, scalable, and open data repository to help pace toward more smart EV knowledge management system. Meanwhile, this work will provide an ontology for 
various EV-related knowledge, which enables rigorous logical interpretation and machine-actionability. With the proposed ontology, EVKG would enable effective integration of critical spatial and semantic information of electric vehicles including the EV charging infrastructures, the electricity transmission network, and the electric vehicle adoptions at different spatial scales. 
 The contribution of this paper can be summarized as follows:
 \begin{enumerate}
     \item To introduce reusability and interoperability, we design an ontology for electric vehicles including Electric Vehicle Adoption Ontology Module, Electric Vehicle Charging Infrastructure Ontology Module, and Electric Transmission Network Ontology Module. 
Instead of reinventing the wheel, We reuse many ontology design patterns to model the spatial, temporal, and semantics aspect of EV data based on GeoSPARQL \citep{battle2011geosparql}, Time Ontology \citep{hobbs2006time}, SOSA SSN Ontology \citep{janowicz2019sosa}, and KnowWhereGraph \citep{janowicz2022know} ontology.
     Figure \ref{fig:ontology} illustrates the overall ontology design patterns.
     
     \item Based on the proposed EV ontology, we construct an electric vehicle knowledge graph called EVKG which includes different geospatial and semantics information about EV data, such as EV charing infrastructures, the electricity transmission network, and the electric vehicle adoptions at different administrative scales. GraphDB\footnote{\url{https://www.ontotext.com/products/graphdb/}} is used as the triple store to support GeoSPARQL-enabled KG queries.
     
     \item We link the constructed EVKG with some existing knowledge graphs such as GNIS-LD \citep{regalia2018gnis}, Wikidata \citep{vrandevcic2014wikidata}, and KnowWhereGraph \citep{janowicz2021knowwheregraph}. For example, instead of redefining the place hierarchy, we reuse the administration region entities \texttt{kwg-ont:ZipCodeArea} and \texttt{kwg-ont:AdministrativeRegion\_3} from KnowWhereGraph which are also linked (\texttt{owl:sameAs}) to Wikidata, GNIS-LD, and DBpedia. 
     Instead of regenerating the transportation network triples, we directly use the national highway planning network subgraph from the KnowWhereGraph. 
\item We propose six competency questions from different EV usage scenarios which are classified into three different question groups to illustrate how we can use EVKG to effectively solve EV and transportation-related questions.
 \end{enumerate}
 
This paper is organized as follows: we discuss some related work in Section \ref{sec:related}. Then, in Section \ref{sec:graph}, we discuss how the EVKG is constructed including the data sources as well as the EVKG ontology design patterns. Next, we demonstrate how we can use EVKG to answer various EV-related competency questions in Section \ref{sec:question}. We conclude this paper in Section \ref{sec:conclude} and discuss the limitations and future works. \section{Related Work} \label{sec:related}

As the electric vehicle sector has become a research hotspot, many studies in the EV sector have been conducted to tackle various issues related to transportation electrification. These studies span a range of subjects, including strategic charging infrastructure placement \citep{dong2014charging, micari2017electric}, E-mobility recommendation \citep{lee2020dc}, transportation equity \citep{hardinghaus2020real}, emergency response \citep{li2022optimal, feng2020can}, and many others \citep{namdeo2014spatial, chen2013locating}. Due to the interconnected nature of the EV sector with other systems, many of these studies require interdisciplinary strategies and the integration of multiple, cross-domain data sources. 
To address this problem, ontologies have been developed to achieve semantic interoperability across different data sources. For instance, \citet{scrocca2021urban} introduced the Urban IoT ontology, which conceptualized the data exchange between service providers and operating IoT devices in the urban area. However, this ontology primarily focuses on the micro-level interactions between EVs and charging infrastructure and does not offer links to external knowledge graph resources or their ontologies. Additionally, the Urban IoT and many other ontology works \citep{santos2018iberian} do not deploy their ontologies for real-world data silos to provide accessible knowledge graph resources. 
In contrast, in our work, we not only provide an ontology to model different aspects of EVs, but also utilize the developed ontology to construct a large-scale EV knowledge graph and connect it to various external knowledge graphs.

Knowledge graphs (KGs) are a novel paradigm for retrieving, reusing, and integrating data from heterogeneous data sources and representing data in a human and machine-readable format \citep{noy2019industry,janowicz2022know}. 
As an important type of knowledge graph, 
 geospatial knowledge graphs (GeoKG) are essentially a symbolic representation of geospatial knowledge. It has become an indispensable component of Symbolic GeoAI \citep{mai2022symbolic} and supports various intelligent geospatial applications such as qualitative spatial reasoning \citep{freksa1991qualitative,zhu2022reasoning,cai2022hyperquaternione}, geographic entity recognition and resolution \citep{alex2015adapting,gritta2018melbourne}, geographic knowledge graph summarization \citep{yan2019spatially},
 geographic question answering \citep{mai2020se,scheider2021geo,mai2021geoqa},  and so on. Nowadays, there are multiple large-scale, open-sourced geospatial knowledge graphs available to use including \textit{GeoNames} \citep{ahlers2013geonames},\textit{ LinkedGeoData} \citep{auer2009linkedgeodata}, \textit{YAGO2} \citep{hoffart2013yago2}, \textit{GNIS-LD} \citep{regalia2018gnis}, and \textit{KnowWhereGraph} \citep{janowicz2021knowwheregraph,janowicz2022know}.

Among all these geospatial knowledge graphs, KnowWhereGraph (KWG) \citep{janowicz2021knowwheregraph,janowicz2022know} is a newly created large-scale cross-domain knowledge graph that integrates datasets at the human-environment interface. It contains various geospatial, demographic, and environmental datasets. Currently, it hosts more than 12 billion information triples which makes it one of the largest geographic knowledge graphs. Please refer to KnowWhereGraph website\footnote{\url{https://www.knowwheregraph.org/}} for a detailed description of the graph. 
Moreover, KWG also provides a collection of geoenrichment services \citep{mai2019deeply,mai2022narrative} and visualization interfaces \citep{liu2022knowledge} on top of the graph which allows the no-expert users to explore, utilize, and analyze graph data seamlessly without any knowledge of Semantic Web. 
The graph also contains co-reference resolution links to multiple existing knowledge graphs including Wikidata \citep{vrandevcic2012wikidata}, GNIS-LD \citep{regalia2018gnis}, GeoNames \citep{ahlers2013geonames}, etc. In this work, instead of regenerating subgraphs to describe place hierarchy and transportation networks, we build co-reference resolution links between EVKG and KWG on the US zip code area entities and
reuse the ontology and subgraph of the national highway planning network. Please refer to Section \ref{sec:source} for a detailed description. \section{Electric Vehicle Knowledge Graph} \label{sec:graph}
\subsection{Data Sources} \label{sec:source}
In this work, we construct the electric vehicle knowledge graph (EVKG) by retrieving, cleaning, integrating, and synthesizing information from various electric vehicle-related public data sources. 
Specifically, we develop the EVKG based on external data sources in terms of four aspects including electric vehicle basic specifications, electric vehicle registration information, electric vehicle charging infrastructure, and electricity transmission networks. Moreover, we also link EVKG with other open-sourced knowledge graphs such as KnowWhereGraph \citep{janowicz2022know}, GNIS-LD \citep{regalia2018gnis} which are also connected with DBpedia \citep{auer2007dbpedia}, Wikidata \citep{vrandevcic2012wikidata}, and GeoNames \citep{ahlers2013geonames}.
Three criteria are used to select EV-related data sources: 1) finer spatial resolution: EV-data should be recorded in small geographic units (e.g., zip code level); 2) finer temporal resolution: EV-data should be updated frequently (e.g., annually); 3) reliable data resources: EV-data should be collected from reliable organizations and institutions (e.g., governments and large NGOs). 
Based on these criteria, we \yanlin{collect} data from the open-source data repositories listed below.

\paragraph{Electric Vehicle Adoption.} 
To collect the electric vehicle's basic specifications (e.g., makes, models, manufacturers, etc.) and track the adoption of electric vehicles annually across different geographic units, we use the electric vehicle registration records amongst local administrative region to indicate snapshots in time of the electric vehicles "on the road". As one of the groundswell support for improving public data accessibility in transportation electrification, the Atlas EV Hub\footnote{\url{https://www.atlasevhub.com/materials/state-ev-registration-data/\#data}} provides a website for electric vehicle registration databases for multiple US states. In this work, we utilize the Atlas EV registration data and aggregate data records to the zip code level to protect user privacy while enabling fine-grained geographic data access. We develop a triplification pipeline to consume the Atlas EV Hub and build the electric vehicle adoption sub-graph. This subgraph will be continuously updated automatically along with the Atlas EV Hub. The Atlas EV Hub collects raw data from vehicle registration agencies, with each row containing a vehicle registration record including the first 8 digits of the Vehicles Identification Number (VIN-8), the corresponding zip code, the model year, and registration valid/expiration data. This could enable policymakers to track the rapidly-evolving conditions of the electric vehicle market without leaking the individual EV registration data since the last six digits of the whole VIN are erased. Meanwhile, with the VIN-8 code and model year provided, we can identify the vehicle information at product levels including vehicle make/model, duty-level category, use case, etc. For each unique type of electric vehicle product, we additionally collect its compatible charger types and connector types from the public repository of EVS pecifications \footnote{\url{https://www.evspecifications.com/}}.

\paragraph{Electric Vehicle Charging Infrastructure.} 
Charging stations are essential infrastructures that safely deliver energy from the electric grid to the battery of an electric vehicle. The U.S. Department of Energy establishes an open-source repository of nationwide electric vehicle charging stations\footnote{\url{https://afdc.energy.gov/fuels/electricity\_locations.html\#/find/nearest?fuel=ELEC}} and provides continuously updated data to the public. We build a triplification pipeline which uses this repository as the main data source to build and continuously update the proposed charging station sub-graph of EVKG. More specifically, each station consists of information regarding its geo-location, charging capacity, usage restrictions, and other contextual attributes. 

\paragraph{Electric Transmission Network.} 
Electric transmission can be seen as a bulk movement of electrical energy from the generating sites such as power plants to electrical substations. The electric transmission network acts as the backbone for the transport of electric power energy across large geographic regions. We integrate the electric transmission network data into our EVKG and enrich them with geographical and contextual information. This aims at facilitating the construction of a macro-grid system with seamless interconnection with the charging infrastructure and the electric vehicle adoption market, which further helps to meet the incoming surge of electric vehicle charging demand. We, therefore, collect the electric power transmission network data across the U.S. from the Homeland Infrastructure Foundation Level Database (HIFLD)\footnote{\url{https://hifld-geoplatform.opendata.arcgis.com/datasets/geoplatform::electric-power-transmission-lines/}}. This repository depicts the main infrastructure components and their interconnections of the electric transmission network, including electric energy transmission lines, power plants, and substations in the U.S. Both the spatial and contextual features in the power sector (e.g., voltage levels, capacities, and install ways) are integrated into our electric transmission network subgraph.

\paragraph{Place Hierarchy and Transportation Network.} Since KWG covers several types of multiscale geospatial reregions which are also essential for electric vehicle management in various place hierarchies (e.g., zip code areas, transportation networks, U.S. cities, counties, states, etc.), we link the constructed EVKG with KnowWhereGraph instead of reinventing the knowledge graph schema and regenerating knowledge graph triples for this information. More concretely, we first build co-reference resolution links (i.e., \texttt{owl:sameAs}) between EVKG and KWG on the U.S. zip code area entities. 
Moreover, as the road network is also a crucial subsystem to encompass in EVKG, we reuse the ontology and subgraph of the national highway planning network from the KWG graph.
There are three advantages of this practice.
First, by co-referencing our zip code entities to those in the KWG, we can skip the process of regenerating all the place hierarchy triples but directly use those from KWG. 
Second, KWG also links its zip code entities to many geographic features such as natural disasters, soil profiles, climate zones, etc. By using the zip code entities as the transit node, we can ask questions across graphs like \textit{which zip code areas had more than 10 electric vehicle charging stations in 2021 but were affected by multiple wildfires in the past. }
Third, since KWG also contains various co-reference resolution links to other knowledge graphs, this practice essentially makes our EVKG become an important component in the Linked Open Data Cloud \citep{assaf2015linkedopendatacloud}.

\subsection{EVKG Ontology} \label{sec:ontology}

While a broad range of domains and application facets are worth encompassing, the EVKG ontology strikes a balance between intricate expressiveness and brief logical statements. In contrast with other already available microscopic ontology suites in the EV space, the EVKG takes a more macroscopic perspective and targets to support the backbone of an integrated ecosystem of the EV-centric environment across geographic regions. Concretely, EVKG models and revolves around the most important classes and relationships of electric vehicle adoption, electric vehicle charging infrastructure availability, electric transmission networks, and major road networks with geo-enrichment.

\begin{figure}[htb]
        \centering
        \includegraphics[width=1.0\textwidth]{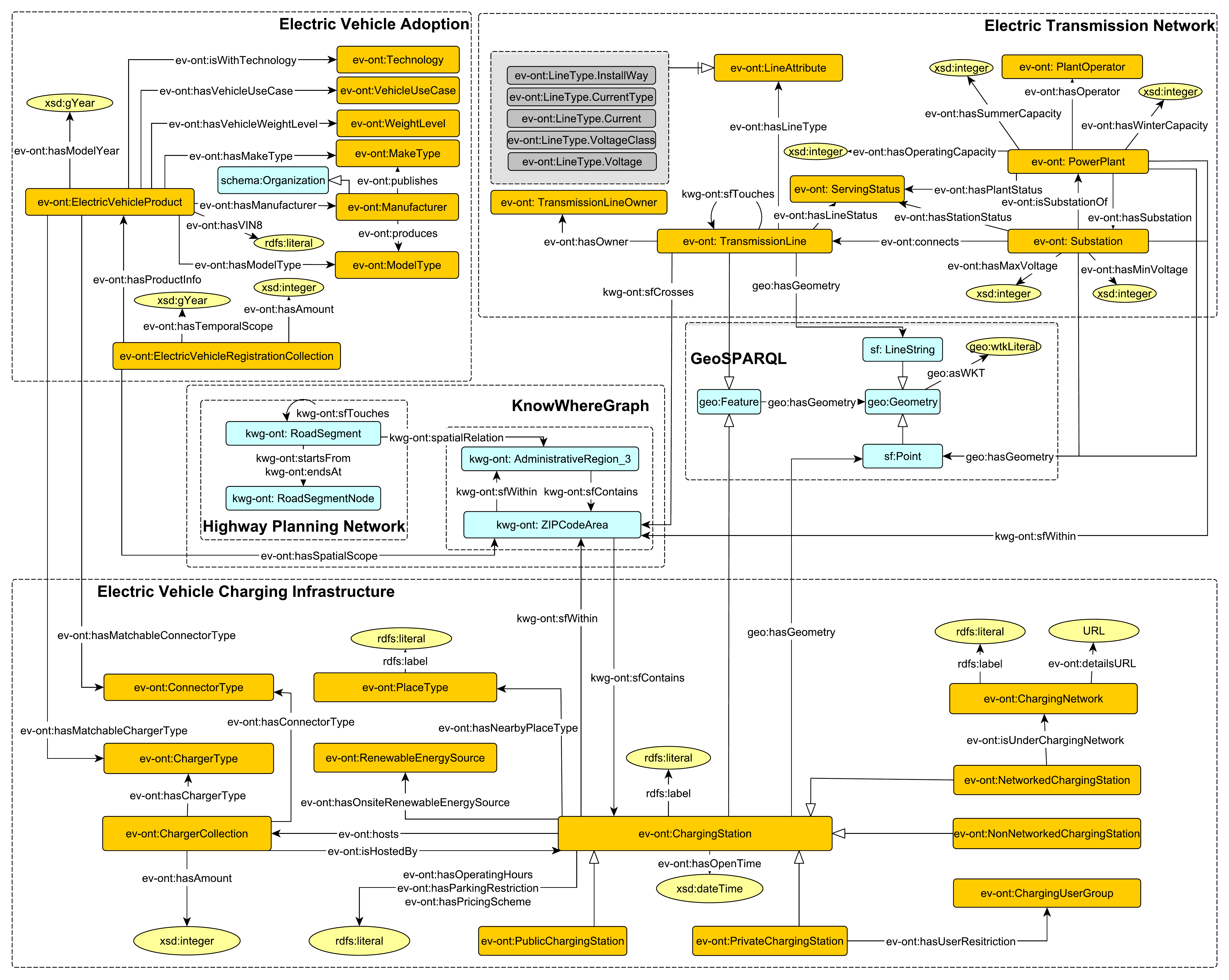}
        \caption{The ontology design of the electric vehicle knowledge graph. 
        Orange nodes indicate classes while yellow nodes indicate literals. Blue nodes indicate these classes are reused from other ontologies including GeoSPARQL, Time Ontology, and KnowWhereGraph ontology.
        White arrows without edge labels refer to the \texttt{rdfs:subClassOf} predicate.
        }
        \label{fig:ontology}
\end{figure}

The proposed EVKG ontology could serve as the backbone for building an ecosystem that represents, retrieves, and integrates these heterogeneous data in the EV domain. By linking the EVKG with external knowledge graphs, we target to further address the bottlenecks of scalability and data silos problems while incorporating diverse data across a wide range of domains outside of the EV domain. The ontology is formally expressed in the W3C-recommended framework and uses the language of RDF\footnote{\url{https://www.w3.org/RDF/}} and OWL\footnote{\url{https://www.w3.org/OWL/}}. Notably, the designed ontology integrates not just disparate data silos, but importantly, also their relationships, in a way that is readable for both humans and machines. The ontology module of each subgraph is visualized in Figure \ref{fig:ontology} and will be discussed as follows.

\subsubsection{Electric Vehicle  Adoption Ontology Module} \label{sec:ev-adoption-ont}
The upper left dashed box in Figure \ref{fig:ontology} illustrates our electric vehicle adoption ontology module.
One critical class in this module is \texttt{ev-ont:ElectricVehicleRegistrationCollection} which indicates the class of a collection of electric vehicle registrations that share the same spatial and temporal scope, as well as product information by using properties: \texttt{ev-ont:hasSpatialScope}, \texttt{ev-ont:hasTemporalScope}, and \texttt{ev-ont:hasProductInfo}. The idea of using a class to denote a collection of EV registration collection is inspired by the \texttt{sosa:ObservationCollection} class proposed by the extensions to the Semantic Sensor Network Ontology \citep{cox2018extensions,zhu2021environmental}. 
In the following, 
we will use an instance \texttt{evr:evregcol\_1} of the \texttt{ev-ont:ElectricVehicleRegistrationCollection} class as an example to demonstrate how to use this module. Here, \texttt{evr:evregcol\_1} represents a specific registration collection of 36 electric vehicle registration records which share the same spatial and temporal scope, as well as product information. First, all these records have the same spatial scope by using property \texttt{ev-ont:hasSpatialScope} -- the ZIP code area 07677 with type \texttt{kwg-ont:ZipCodeArea}. Second, they have the same temporal scope (\texttt{ev-ont:hasTemporalScope}) of the year 2019 which are denoted as \texttt{``2019''\^{}\^{}xsd:gYear}. Third, this collection has the same electric vehicle product information of the type \texttt{ev-ont:ElectricVehicleProduct} by using property \texttt{ev-ont:hasProductInfo}. This product information has a maker type \texttt{evr:BMW} which is an instance of \texttt{ev-ont:MakeType}. It also has a product type -- i3 which is an instance of \texttt{ev-ont:ModelType} which has a model year \texttt{``2018''\^{}\^{}xsd:gYear} by using the property \texttt{ev-ont:hasModelYear}. We also use \texttt{ev-ont:isWithTechnology} to associate each product information entity with its technology type, e.g., battery electric vehicle of type \texttt{ev-ont:Technology}. The product information entity also has a \texttt{ev-ont:Manufacturer} -- BMW of North America Inc., a \texttt{ev-ont:VehicleUseCase} of compact, and a \texttt{ev-ont:WeightLevel} of light-duty vehicles whose weight level is less than 6,000 lbs. This product information entity also specifies the charger type (\texttt{ev-ont:ChargerType}) -- Level 2 and DC fast charging, and connector type (\texttt{ev-ont:ConnectorType}) -- J1772 connector and CCS connector.
Here, EVKG defines two properties of \texttt{ev-ont:hasMatachableChargerType} and \texttt{ev-ont:hasMatachableConnectorType} to link the charging specification differences with respect to vehicle makes and models to the charging infrastructure.

\subsubsection{Electric Vehicle  Charging Infrastructure Ontology Module} \label{sec:ev-charge-station-ont}

The lower part of Figure \ref{fig:ontology} shows our Electric Vehicle Charging Infrastructure Ontology Module.  
As the key class in this module, \texttt{ev-ont:ChargingStation} indicates a class of units of electric vehicle charging infrastructure for refueling electric vehicles. 
In EVKG, we model \texttt{ev-ont:ChargingStation} as a subclass of \texttt{geo:Feature} which is defined by the GeoSPARQL ontology \citep{battle2011geosparql}. So each charging station will have a \texttt{sf:Point} type node to indicate its spatial footprint. Moreover, we explicitly materialize the spatial relations between a charging station and a zip code area (\texttt{kwg-ont:ZipCodeArea}) as knowledge graph triples by reusing the spatial relation properties defined by KnowWhereGraph -- \texttt{kwg-ont:sfWithIn} and \texttt{kwg-ont:sfContains}. The basic attributes of a charging station such as its opening time, operating hours, parking restrictions, pricing scheme are simply modeled by dedicated properties: \texttt{ev-ont:hasOpenTime}, \texttt{ev-ont:hasOperatingTime}, \texttt{ev-ont:hasParkingRestriction}, and \texttt{ev-ont:hasPricingScheme}.

A charging station may host multiple chargers and some of them could share the same characteristics. Chargers may have different charging levels and connector types since different electric vehicles have different matchable connector types and not all chargers offer all types of EVs. However, given a set of chargers that are provided by the same charging station and share the same characteristics, generating assertions for each of them could be bothersome and cause unnecessary redundancy. 
To strike the balance between triple store storage efficiency and semantic accuracy, EVKG defines the concept of \texttt{ ev-ont:ChargerCollection} to model a set of chargers provided by one charging station that belongs to the same \texttt{ ev-ont:ChargerType} and with the same \texttt{ ev-ont:ConnectorType}.  
The number of chargers contained in an \texttt{ ev-ont:ChargerCollection} is expressed by using the \texttt{ev-ont:hasAmount} property. Notably, EVKG only distinguishes different \texttt{ev-ont:ChargerType} by using the charging levels that represent their charging speeds, not based on different brands of them. In the U.S. market, electric vehicle supply equipment (EVSE) comes in three levels: Level 1, Level 2, and Level 3. While Level 1 and Level 2 alternating-current (AC) chargers require long hours to recharge the battery to its full capacity, direct-current fast chargers (DCFCs) can recharge a battery by 80\% in a 15-minute charge session which reduces charging waiting time and indirectly decreases the demand for a larger number of charging stations. Moreover, unlike gas stations, there is no universal connector type (i.e., the socket through which you connect a charger to the car) for EV charging. For instance, J1772COMBO (Combined Charging System, CCS), CHAdeMO, and TESLA are three widely accepted connector standards in the U.S. Therefore, EVKG models \texttt{ev-ont:ConnectorType} and \texttt{ev-ont:ChargerType} for each \texttt{ev-ont:ChargerCollection}.

It is also important to distinguish between different types of \texttt{ev-ont:ChargingStation}.
As the user access permission of an \texttt{ev-ont:ChargingStation} determines its role in providing public services, EVKG distinguishes between \texttt{ev-ont:PublicChargingStation} and \texttt{ev-ont:PrivateChargingStation}. The former features in its availability to the general public, including shopping center parking, on-street parking, and non-reserved multi-family parking lots, etc., while the latter specifies a charging station that provides exclusive services for designated user groups (\texttt{ev-ont:ChargingUserGroup}) (e.g., designated employee parking). Connecting a charging station to be part of a charging network (\texttt{ev-ont:ChargingNetwork}) improves its online accessibility to EV users and enables its ability to set pricing of EV charging or resell the electricity. An \texttt{ev-ont:ChargingNetwork} means a data management system deployed on electric vehicle supply equipment and connected via an online connection. EVKG defines this type of charging station as \texttt{ev-ont:NetworkedChargingStation}. Each \texttt{ev-ont:NetworkedChargingStation} is connected to its \texttt{ev-ont:ChargingNetwork} using the \texttt{ev-ont:isUnderChargingNetwork} property. In comparison, EVKG uses the concept of \texttt{ev-ont:NonNetworkedChargingStation} to represent the non-networked station that is a stand-alone unit without access control and not available online. The outlined subclasses of \texttt{ev-ont:ChargingStation} are linked to it via \texttt{rdfs:subClassOf} relation.

\subsubsection{Electric Transmission Network Ontology Module} \label{sec:transline-ont}
The upper right dashed box in Figure \ref{fig:ontology} contains our Electric Transmission Network Ontology Module which encompasses the important concepts and roles of the power transmission infrastructure defined in the EVKG. 
Unlike some ontologies in the power system domain that provide detailed descriptions of the grid assets \citep{huang2015knowledge}, EVKG focuses on enabling the geo-enrichment service \citep{mai2019deeply} and cross-domain integrations with other EVKG's subgraphs. In EVKG, the core components of a typical transmission network are conceptualized as \texttt{geo:Feature}, which includes the power generator (\texttt{ev-ont:PowerPlant}) where the electricity is produced, the transmission line (\texttt{ev-ont:TransmissionLine}) that carries electricity over long distances, and the transmission substation (\texttt{ev-ont:Substation}) that connects two or more transmission lines. Each of them is associated with their spatial footprints with type \texttt{geo:Geometry}.

In addition to geometric and topological properties, the developed ontology module introduces a list of data type properties to describe the attributes of each power transmission infrastructure component. First, the electricity capacities of \texttt{ev-ont:PowerPlant} and voltage capacities of \texttt{ev-ont:Substation} are defined as data type properties -- \texttt{ev-ont:hasSummerCapacity}, \texttt{ev-ont:hasWinterCapacity}, \texttt{ev-ont:hasOperatingCapacity}, and \texttt{ev-ont:hasMinVoltage}, \texttt{ev-ont:hasMaxVoltage} respectively.
The serving status of transmission lines, powerplants, and substations are defined as \texttt{ev-ont:ServingStaus} by using object type properties -- \texttt{ev-ont:hasLineStatus}, \texttt{ev-ont:hasPlantStatus}, and \texttt{ev-ont:hasStationStatus} correspondingly.
The attributes of transmission lines such as voltage class, current types, install way, etc. are modeled as object-type properties.
All attribute nodes are entities of type \texttt{ev-ont:LineAttribute}. The reason of using object type properties instead of datatype properties is that those line attributes contain more complex and type-level information. For example, \texttt{ev-ont:VoltageClass} indicates various voltage levels of transmission lines. We model the owner of transmission lines as type \texttt{ev-ont:TransmissionLineOwner}.

\section{Result and Case Study} \label{sec:question}
\subsection{Statistics about Electric-Vehicle-Knowledge-Graph}
At the date of submitting this paper (January 2023), EVKG consists of 83 classes and 69 properties that describe the core concepts in the EV sector. Based on the collected data, more than 27 millions statements are currently included in the EVKG. The entities used for describing the road network are directly collected from KnowWhereGraph, and are directly encapsulated into our EVKG. More detailed statistics can be found in Table \ref{tab1}. Moreover, EVKG will be updated regularly by including newly installed charging stations, incoming EV registrations, new electric vehicle products, upgraded transmission networks, and others. 
The ontology and knowledge graph of EVKG are available here\footnote{\url{https://github.com/EVKG/evkg}}.

\begin{table}[!ht]
\caption{\textbf{Statistics of the electric vehicle knowledge graph (EVKG)}}
\resizebox{\textwidth}{!}{
\begin{tabular}{|cccccccccc|}
\hline
\multicolumn{1}{|c|}{Key class}          & \multicolumn{1}{c|}{\begin{tabular}[c]{@{}c@{}}Charging\\ Station\end{tabular}} & \multicolumn{1}{c|}{\begin{tabular}[c]{@{}c@{}}Charger\\ Collection\end{tabular}} & \multicolumn{1}{c|}{\begin{tabular}[c]{@{}c@{}}ElectricVehicle\\ RegistrationCollection\end{tabular}} & \multicolumn{1}{c|}{\begin{tabular}[c]{@{}c@{}}ElectricVehicle\\ Product\end{tabular}} & \multicolumn{1}{c|}{\begin{tabular}[c]{@{}c@{}}Transmission\\ Line\end{tabular}} & \multicolumn{1}{c|}{Substation} & \multicolumn{1}{c|}{\begin{tabular}[c]{@{}c@{}}Power\\ Plant\end{tabular}} & \multicolumn{1}{c|}{\begin{tabular}[c]{@{}c@{}}Road\\ Segment\end{tabular}} & \begin{tabular}[c]{@{}c@{}}RoadSegment\\ Node\end{tabular} \\ \hline
\multicolumn{1}{|c|}{Number of Entities} & \multicolumn{1}{c|}{50, 143}                                                    & \multicolumn{1}{c|}{52, 370}                                                      & \multicolumn{1}{c|}{429, 682}                                                                         & \multicolumn{1}{c|}{10,602}                                                            & \multicolumn{1}{c|}{93, 047}                                                     & \multicolumn{1}{c|}{75, 327}    & \multicolumn{1}{c|}{12, 556}                                               & \multicolumn{1}{c|}{538, 014}                                               & 1, 076, 028                                                \\ \hline
\multicolumn{10}{|c|}{Total number of statements:27, 608, 442}                                                                                                                                                                                                                                                                                                                                                                                                                                                                                                                                                                                                                                                                                               \\
\multicolumn{10}{|c|}{Total number of entities: 4, 298, 217}                                                                                                                                                                                                                                                                                                                                                                                                                                                                                                                                                                                                                                                                                                 \\
\multicolumn{10}{|c|}{Total number of properties: 69}                                                                                                                                                                                                                                                                                                                                                                                                                                                                                                                                                                                                                                                                                                        \\
\multicolumn{10}{|c|}{Total number of classes:83}                                                                                                                                                                                                                                                                                                                                                                                                                                                                                                                                                                                                                                                                                                            \\ \hline
\end{tabular}}
\label{tab1}
\end{table}

\subsection{Competency Questions}
One strength of the presented EVKG is that it is highly versatile, allowing for answering various competency questions focusing on different domains and use cases. These competency questions can be answered by executing SPARQL queries\footnote{\url{https://www.w3.org/TR/rdf-sparql-query/}}, a standard query language for RDF-based knowledge graphs, that span multiple topics of interest under the EV theme, making a wide range of information readily accessible to users. 

In this section, we discuss multiple exemplary competency questions together with the corresponding queries to showcase the intended use of the designed EVKG ontology and its associated EVKG resource. These examples highlight the diverse range of possibilities offered by the graph from semantic and geospatial queries to cross-domain queries that are expected to be answered based on information across different data silos. These queries are grouped into three main categories. Noted that the contents inside the brackets of each competency question indicate property, specific entities, and classes which can be easily replaced with similar properties, entities, or classes. The discussed queries can be found in the published Github repository and the graph endpoint, which can be executed with ease.

\paragraph {Group 1: Semantic and Geospatial Questions}

\paragraph{Q1. Semantic Questions} \textit{Which [electric vehicle products] have charging cables that match the [CHADeMO connector type]? }

Unlike the simple and universal fuel fill inlet found on fuel vehicles, not all electric vehicles are manufactured with the same charging capabilities. The variety of connector types and charging cable standards across different EV models make it difficult to identify the charging capability of different electric vehicle products. This complexity can be further compounded by factors such as weight, cost, and space limitations, which may affect the acceptance rate and type of on-board fast charging cables for specific EV models. Such a complicated situation makes it confusing for identifying the charging capability of different electric vehicle products.  Fortunately, EVKG can provide a solution to this confusion by providing quick and easy access to knowledge about the semantic descriptions of different electric vehicle products through SPARQL queries (\yanlin{see Listing \ref{q:q1} in Appendix \ref{sec:query}}).

\paragraph{Q2. Geospatial Questions} \textit{Which [charging stations/road segments/transmission lines/power plants/ substations] are [located in/pass through] [King county]?}

To answer the above geospatial questions in a typical GIS environment, we need to collect geospatial datasets from various domains, clean them, and reproject them into a shared proper  geographic/projection coordinate system as various GIS layers. And then we can write spatial queries across different GIS layers to answer these questions. 
In contrast, EVKG can answer these geospatial questions with simple geospatial SPARQL queries (\yanlin{see Listing \ref{q:q2}  in Appendix \ref{sec:query}}). 
EVKG contains various types of geospatial features. As described in Section \ref{sec:ontology}, we use GeoSAPARQL ontology to encode each geospatial feature's spatial footprints and use KnowWhereGraph's spatial relation properties (e.g., \texttt{kwg-ont:sfWithin}) to explicitly generate some spatial relation triples to speed up the geospatial queries.
So the above geospatial questions can be answered by either using SPARQL queries with GeoSPARQL functions such as \texttt{geo:sfContains} which explicitly compute spatial relations based on the stored spatial footprints of geospatial entities, or using the prematerialized spatial relation properties such as \texttt{kwg-ont:sfContains}. The drawback of using the first approach is that, as shown by multiple studies \citep{regalia2019computing,mai2021geoqa,mai2022towards}, answering topological questions by explicitly computing the spatial relations among the stored geometries will suffer from sliver polygon problem which might lead to wrong answers. For example, as shown in \citet{mai2022towards}, \texttt{dbr:Seal Beach,\_California} intersects \texttt{dbr:Orange\_County,\_California} based on their OpenStreetMap polygon geometries although we know the former should locate inside of the later one. By using the the prematerialized spatial relation triples, we can avoid this kind of problems.

\paragraph{Q3. Semantic and Geospatial Questions}\textit{Which and where are the [public charging stations] operating ["24 hours daily"] that a [Nissan Leaf 2021] vehicle with a membership of the [ChargePoint] network can use for [fast charging] within ZIP code [95814]?}

\begin{figure}[htb]
        \centering
        \includegraphics[width=0.6\textwidth]{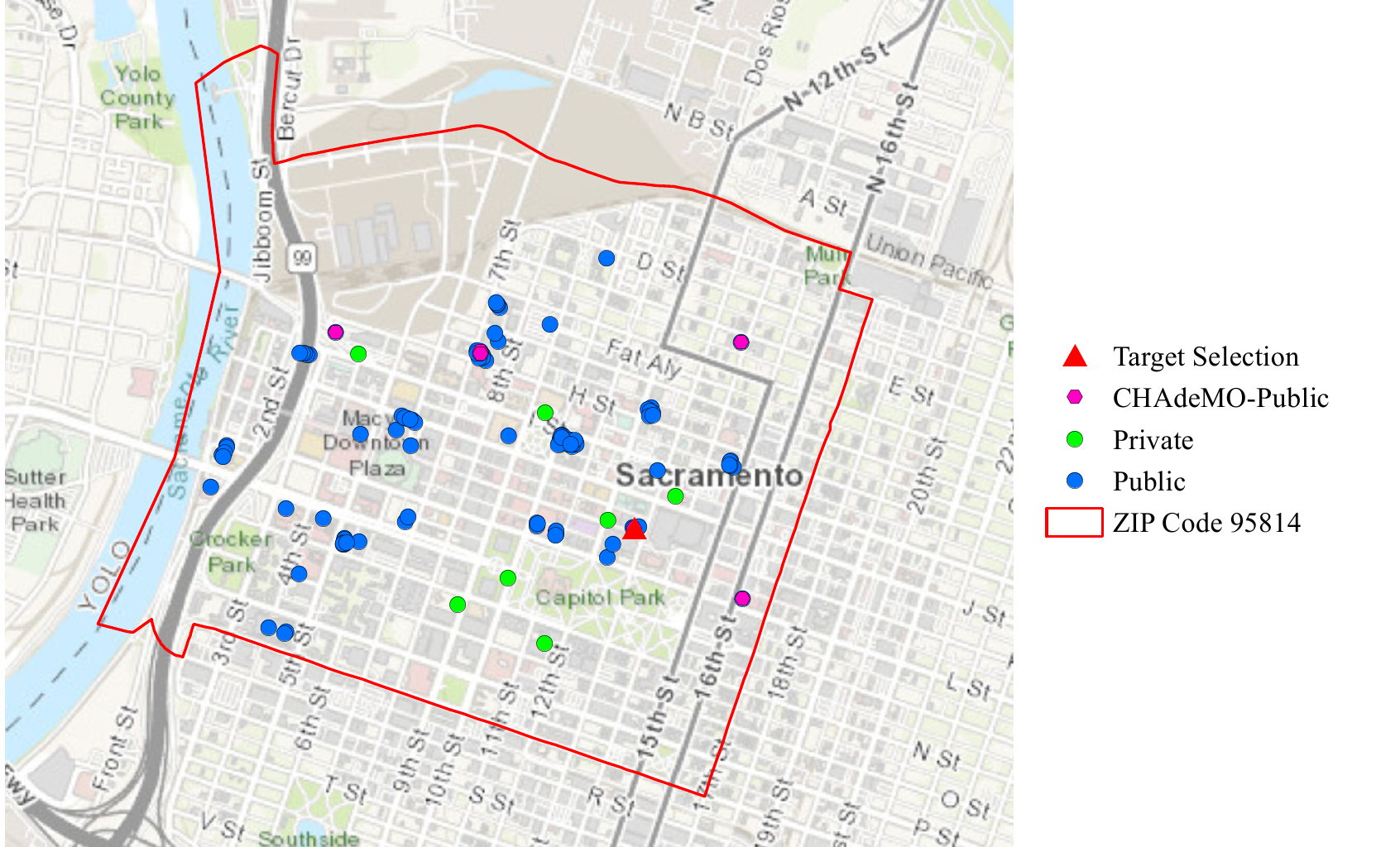}
        \caption{The illustration of the charging station distribution with semantic conditions within the ZIP code area 95814 in Sacramento City. The red triangular represents the target selection that satisfies all the geospatial and semantic conditions defined by Competency Question Q3.}
        \label{fig:Q3}
\end{figure}

The EVKG can serve as a search engine for personalized E-mobility with its rich semantic and geospatial resources. Compared to the general fuel station retrieval services, charging station retrieval encounters more variety in semantic preferences. Often, when people search for charging stations, they consider both geospatial locations and other specific characteristics. For instance, EV drivers searching for charging may consider the user restrictions, charging network membership requirements, and connector types, in addition to the distance and charging speed of a station. The answer of Q3 obtained from EVKG (\yanlin{see Listing \ref{q:q3} in Appendix \ref{sec:query}}) is visualized as the target selection in Figure \ref{fig:Q3}.  

\paragraph{Group 2: Spatial and Temporal Aggregation Questions}

\paragraph{}The EVKG offers a powerful and versatile platform for quantitatively understanding the complex interplay between different types of EV chargers, geospatial regions, and EV adoptions. It allows for a wide range of data analysis and aggregation. Questions 4-5 are two examples presenting a through examination of the distribution of electric vehicle adoption and charging infrastructure, distinguished by the utilized types of connectors over time and space. These examples serve as a demonstration to showcase the extensive capabilities of EVKG in delivering an all-encompassing and comprehensive overview of the electric vehicle theme.

Acting as the crucial coupling between electric vehicles and charging equipment, connectors play a critical role in the seamless transmission of power. The strategic placement and design of charging infrastructure equipped with proper types of connectors can greatly impact the growth and evolution of the automotive industry, as well as the efficiency and stability of power grid systems. In short, connectors are essential building blocks that impact the future of sustainable transportation and energy. Therefore, not only should the EV charging infrastructure match the EV adoption (demand) in terms of the total numbers, but it should also provide a tailored level of chargers with matchable connector types to ensure compatibility with different types of electric vehicles. Otherwise, if there is overproduction and sale of electric vehicles with a specific connector type, but insufficient infrastructure support, it would result in a critical shortage of public charging resources. Similarly, if there is an excessive deployment of charging infrastructure for that particular connector type, it would cause a waste of investment and resources. This highlights the importance of carefully considering and balancing the production and infrastructure to ensure optimal utilization and long-term success of the EV market. A comprehensive and inclusive approach should encompass the variations in both the spatial and temporal dimensions of these elements. From a temporal perspective, the deployment of charging stations, with their varying numbers of EVSEs, as well as the adoption of electric vehicles, have undergone significant evolution over time. When considering the spatial dimension, there is a noticeable heterogeneity across various local regions. Understanding these changes and distributions can provide valuable insights for policymakers and planners, helping them to make informed decisions about public funding investments and improve policies accordingly. 

\paragraph{Q4. Temporal Aggregation Questions} \textit{How does the fast charging resource of the [CCS], [CHAdeMO], and [TESLA] types per matchable electric vehicle evolve over the temporal scope in [New Jersey]?}

\begin{figure}[htb]
        \centering
        \includegraphics[width=0.8\textwidth]{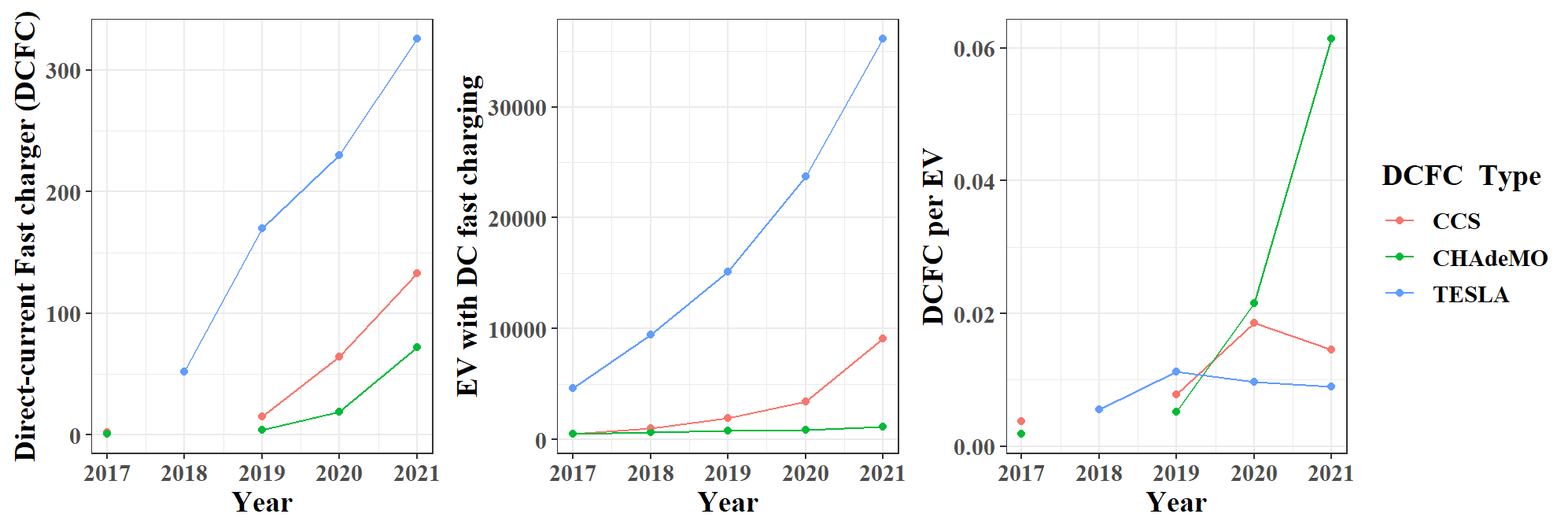}
        \caption{The annual variation trend of the EV registration numbers, direct-current fast charging (DCFC) infrastructure numbers, and average charger share per registered EV across the entire state of New Jersey from 2017 to 2021, regarding different types of connectors. For the years in which no data is presented, there are gaps in the raw data source that has yet to be filled. These three figures are visualizations of the results of the SPARQL query that answer Competency Question Q4.}
        \label{fig:temporal}
\end{figure}

While the J1772 has become the universal standard for Level 2 charging in North America, the variety of connector types for fast charging may demand a more comprehensive understanding to evaluate the markets of EV adoptions and charging infrastructures. By querying the direct-current fast chargers (DCFCs) and grouping them according to connector categories, temporal scopes, and administrative levels such as by state (\yanlin{see Listing \ref{q:q4a}  in Appendix \ref{sec:query}}), it is possible to gain valuable insights into the current state of different types of EVs and corresponding fast charging infrastructure. Additionally, by comparing the number of DCFCs to the number of registered electric vehicles (\yanlin{see Listing \ref{q:q4b}  in Appendix \ref{sec:query}}), we are able to calculate the DCFC resource per registered EV (\yanlin{see Listing \ref{q:q4c}  in Appendix \ref{sec:query}}), providing a clear picture of the fast charging availability for different types of electric vehicles.

As seen in Figure \ref{fig:temporal}, our EVKG data show that both the number of electric vehicles and the number of fast charging infrastructures in the New Jersey state continue to grow during the years with data provided. However, it also highlights that the availability of public fast charging for specific types of vehicles, including Tesla and vehicles applicable to the CCS connector, may not be increasing as quickly as their adoption rates. This is likely due to the deficiency in constructing these specific types of charging stations, which is not keeping pace with the surging demand for them. By breaking down data silos of EV registration records and charging infrastructure repositories, our EVKG is able to provide an integrated view of these two domains, and give a more consolidated and accurate understanding of such questions.

\paragraph{Q5. Spatial Aggregation Queries} \textit{How many registered electric vehicles equipped with the [CCS] type connector are there in each [ZIP code areas] of [New Jersey] in [2021]? How many [CCS chargers] are there in those [zip code areas]? What about the CCS Charger per EV with CCS-type connectors in each [zip code area]?
}

\begin{figure}
     \centering
     \begin{subfigure}[b]{0.48\textwidth}
         \centering
         \includegraphics[width=\textwidth]{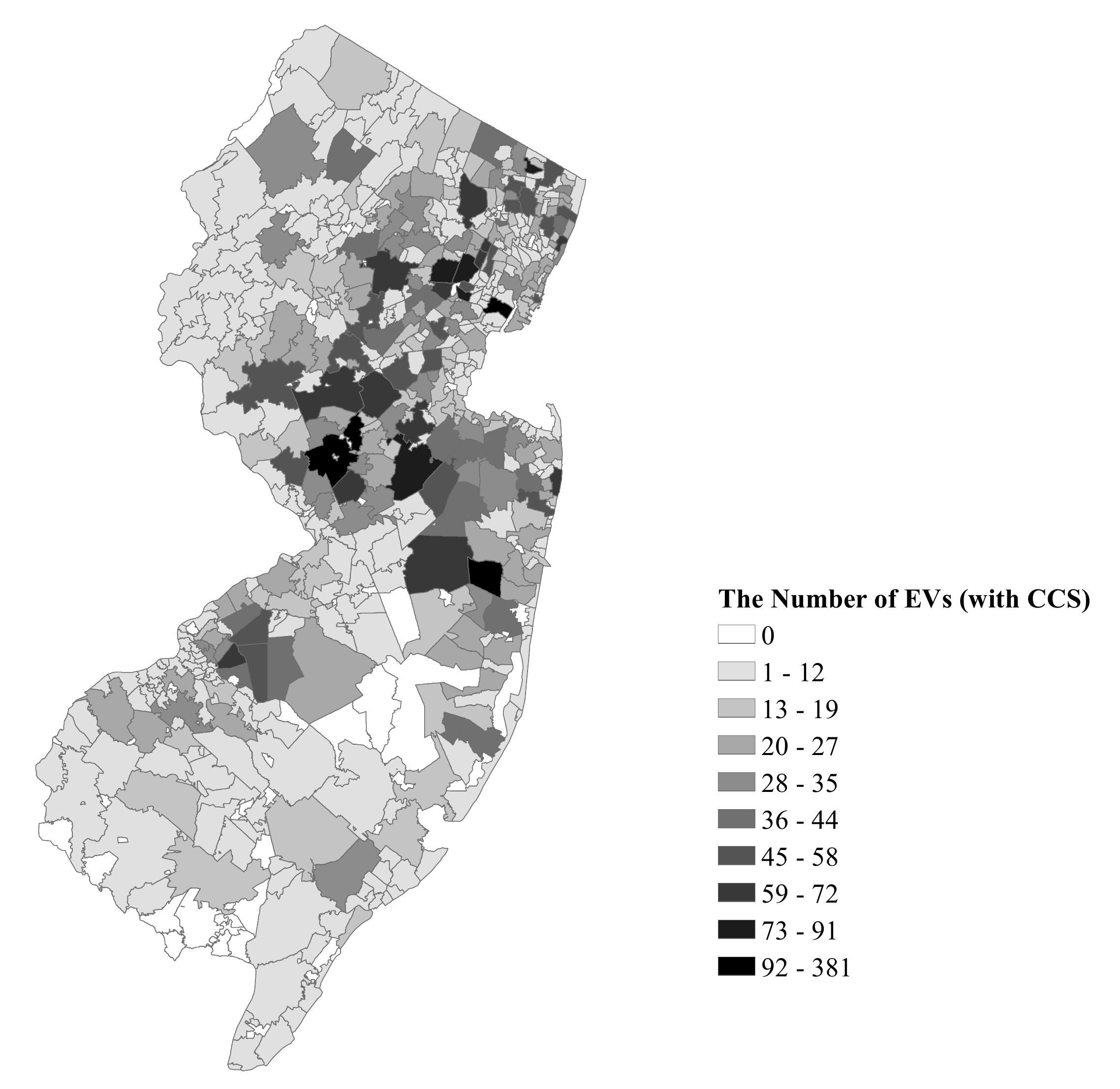}
         \caption{The distribution of CCS-EV registrations}
         \label{fig:ev_adoption}
     \end{subfigure}
     \hfill
     \begin{subfigure}[b]{0.48\textwidth}
         \centering
         \includegraphics[width=\textwidth]{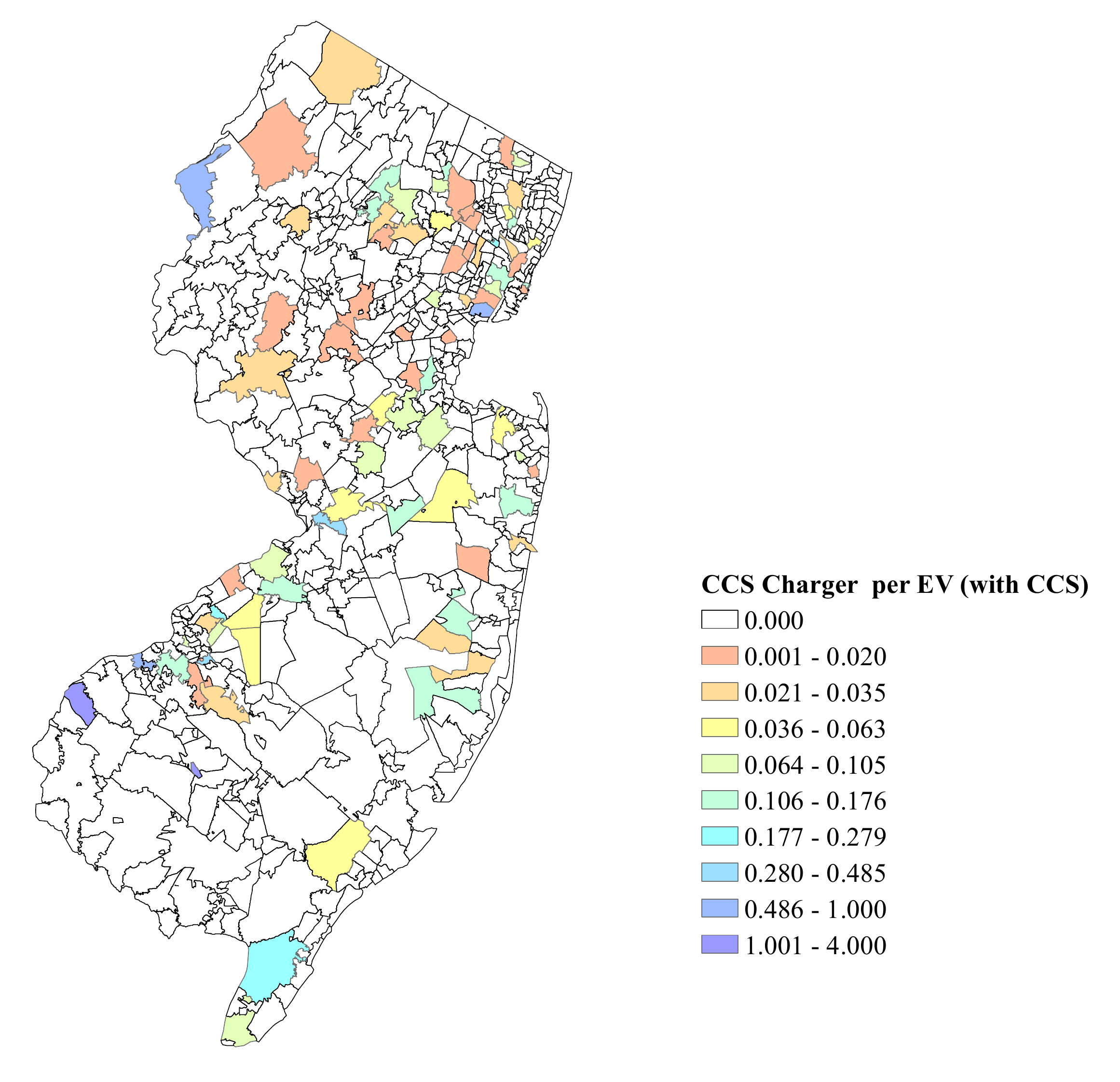}
         \caption{The average CCS-charger share per CCS-EV}
         \label{fig:ccs_charger_per_ev}
     \end{subfigure}
     \caption{ The spatial distributions of the CCS-type electric vehicle registrations and average charging infrastructure share aggregated at zip code level across the New Jersey state in 2021. Figure \ref{fig:ev_adoption} and \ref{fig:ccs_charger_per_ev} are visualizations of the results of the SPARQL queries that answer Competency Question Q5.}
\end{figure}

Public charging stations are not as ubiquitous as gas stations. This disparity addresses the importance of evaluating the spatial distribution of EV registrations and charging resources to effectively plan for charging infrastructure at a more granular level. Among the various types of fast charging options available, the CCS connector market is particularly complex. While companies like Tesla have established their own charging network exclusively for their vehicles, and there are a limited number of EVs compatible with CHAdeMO connectors, the CCS standard is supported by a diverse range of charge point operators and automobile manufacturers. This complexity adds an extra layer of consideration for those involved in charging infrastructure planning and deployment. \yanlin{With just simple queries (\yanlin{see Listing \ref{q:q5a}-\ref{q:q5b}  in Appendix \ref{sec:query}})}, our EVKG enables end users of easy access to analyzing spatial information about targeted elements at a highly specific ZIP code level. Figure \ref{fig:ev_adoption} and \ref{fig:ccs_charger_per_ev} visualize the results of Q5.

\paragraph{Group 3: Cross-Domain Questions}

\paragraph{Q6. Cross-Domain Complex Questions} \textit{ Which zip code areas in the New Jersey State with significant charging resource shortage can potentially take advantage of the high-voltage transmission lines that pass through for installing the direct electricity source for DCFC stations?}

\begin{figure}[htb]
        \centering
        \includegraphics[width=0.5\textwidth]{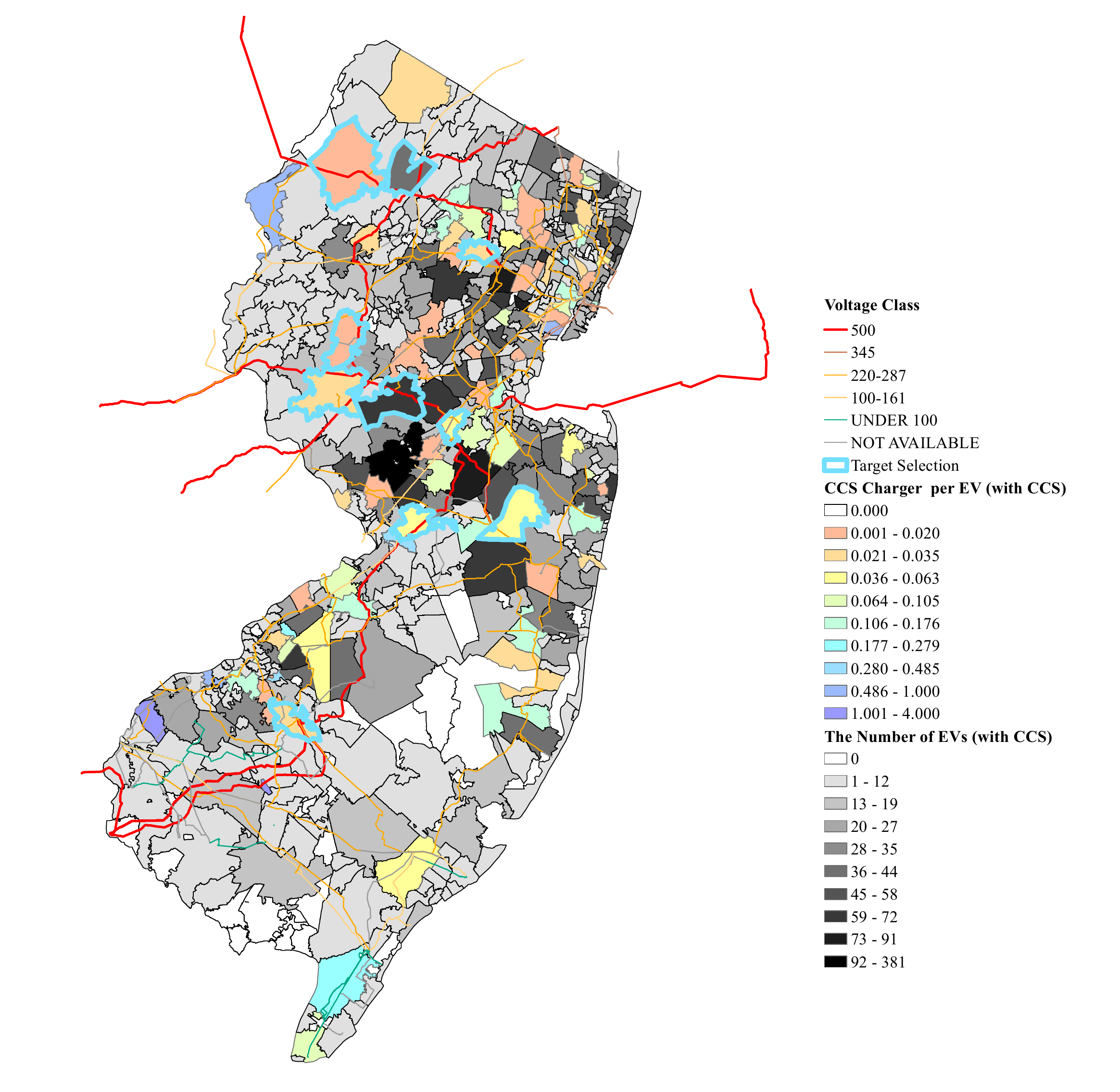}
        \caption{The distribution of the zip code areas in New Jersey which satisfy the requirement specified in Competency Question Q6. The boundaries of the selected zip code areas are highlighted in bright blue.}
        \label{fig:EV_spatial}
\end{figure}

The advert impacts of EV charging stations loads on the voltage profile of distribution networks have been studied by a number of researchers\citep{geske2010modeling, juanuwattanakul2011identification, zhang2016research}. Typically, the fast charging stations could overload the distribution system when connecting to the distribution side.The immediate connection/disconnection of EV loads may also cause power quality issues. Therefore, fast/super-fast charging stations are best fed from transmission lines \citep{rahman2020cascaded}. The placement of EV charging stations may also need to be prioritized in areas that experience a significant shortage of charging resources from the equity perspective.
From a viewpoint of charging infrastructure planning, our EVKG can offer informative insights to enhance the strategic planning of charging station placement.
To achieve this, we can further take the advantage of our EVKG to answer this complex question. Based on the results in Q5, and the fact that the available DC fast chargers require electricity inputs of at least 480 volts, we \yanlin{can query} the EVKG SPARQL endpoint to retrieve the zip code areas that are passed through by such transmission lines above this voltage threshold, in addition to the constraint of low charging resource share and a larger number of registered EVs (\yanlin{see Listing \ref{q:q6a}-\ref{q:q6b}  in Appendix \ref{sec:query}}). The target zip code areas are illustrated in Figure \ref{fig:EV_spatial}. As this is an exemplary answer, the result may be further improved by more well-defined considerations. 

 \section{Conclusion and Discussion} \label{sec:conclude}
While an emphatic global shift to vehicle electrification is undergone, there are still factors impeding its success. Inadequate data sharing and integration can be among the top factors. In this work, \yanlin{the EVKG is developed to serve} as a comprehensive knowledge management to support more efficient EV charging and infrastructure planning. The ontology integrates critical aspects of EVs including EV adoption, EV charging infrastructure, and electricity transmission network. \yanlin{The EVKG is complemented with} several EV-related subgraphs from KnowWhereGraph that are about transportation networks and place hierarchy.

\yanlin{The development of the EVKG includes identifying core modules in the EV industry and collecting data repositories that represent these aspects. An ontology is designed to model these key concepts and their logical relationships, which are encapsulated into different ontology modules. The EVKG contextualizes and enriches the data sources into a united graph structure format, making it efficient to consolidate and query. The strengths of the EVKG are demonstrated through several exemplary competency questions.} To this end, we are able to provide an integration framework that can efficiently support smart EV knowledge management and make an extensive range of EV applications effortless.

\yanlin{Despite only selected data sources being included in the current version of EVKG, the developed ontology enables the further incorporation of other data resources.} With existing open KGs and more data silos to be identified, the EVKG would grow continuously so as to work more remarkably in the future. For instance, the constructed EVKG \yanlin{can} serve as the backbone of E-mobility. With adequate information integration of the geo-location and the semantic annotations of the EVSE infrastructure and EV models, decision-making on matching EV charging stations would be much more efficient. Moreover, with its scalable nature, the EV-charging-centric knowledge graph would enable convenient linking and integration with existing open geo-enrichment service stacks \citep{mai2019deeply,mai2022narrative,janowicz2022know}. With more use case-focused data for additional environmental disasters (e.g., floods or wildfires) or urban planning data, a wide range of knowledge reasoning and discovery can be conducted downstream. This work also makes EV-related data-driven decision-making and data analytics substantially more effective, accessible, and affordable. 

In the future, we plan to further integrate EVKG with other open knowledge graph repositories, such as Wikidata and GeoNames, so that more socioeconomic and environmental challenges beyond electric transportation can be comprehensively addressed. Furthermore, a set of shape constraints can be developed to validate the incoming data to EVKG so as to improve the data quality \citep{zhu2021sosa}. Last but not least, new analytical methods that can work on knowledge graphs could be studied to advance the discovery of insights from a multidisciplinary context.

\addcontentsline{toc}{section}{References}

\bibliographystyle{havard}

\newpage
\section{Appendix}

\subsection{SPARQL Queries} \label{sec:query}

\begin{lstlisting}[
language=SPARQL,
captionpos=b, 
caption={\yanlin{Query for electric vehicle products have charging cables that match the CHADeMO connector type.}}, 
label={q:q1},
]
SELECT DISTINCT ?lev WHERE { 
	?ev a ev-ont:ElectricVehicleProduct.
    	?ev ev-ont:hasMatchableConnectorType evr:connectortype.CHAdeMO.
    	?ev rdfs:label ?lev.
}
\end{lstlisting}

\begin{lstlisting}[
language=SPARQL,
captionpos=b, 
caption={\yanlin{Query for retrieving the charging stations/road segments/transmission lines/power plants/ substations located in/pass through King county. We use ``UNION'' here to denote that those clauses are alternatives to collect different types of geospatial features. }}, 
label={q:q2},
]
SELECT * WHERE { 
    ?county a kwg-ont:AdministrativeRegion_3.
    ?county rdfs:label "King".
    ?zipcode a kwg-ont:ZipCodeArea.
    ?zipcode kwg-ont:sfWithin ?county.
    {?road a kwg-ont:RoadSegment.
    ?road kwg-ont:sfWithin ?county.}
    UNION
    {?transline a ev-ont:TransmissionLine.
    ?transline kwg-ont:sfCrosses ?zipcode.}
    UNION
    {?char_station a ev-ont:ChargingStation.
    ?char_station kwg-ont:sfWithin ?zipcode.}
    UNION
    {?substation a ev-ont:Substation.
    ?substation kwg-ont:sfWithin ?zipcode.}
    UNION
    {?powerplant a ev-ont:PowerPlant.
    ?powerplant kwg-ont:sfWithin ?zipcode.}
    }
\end{lstlisting}

\begin{lstlisting}[
language=SPARQL,
captionpos=b, 
caption={\yanlin{Query for the information of the public charging stations operating 24 hours daily that a Nissan Leaf 2021 vehicle with a membership of the Charge-Point network can use for fast charging within ZIP code 95814.}}, 
label={q:q3},
]
SELECT DISTINCT ?co ?station ?sWKT
WHERE {{    
    ?zipcode a kwg-ont:ZipCodeArea.
    ?zipcode rdfs:label "zip code 95814".
    ?station a ev-ont:PublicChargingStation.
    ?station kwg-ont:sfWithin ?zipcode.
    ?station ev-ont:hosts ?chargerCollection. 
    ?chargerCollection ev-ont:hasConnectorType ?co.
    ?station ev-ont:hasOperatingHours "24 hours daily  ".
    ?station ev-ont:isUnderChargingNetwork 
        evr:chargingnetwork.ChargePointNetwork.
    ?station geo:hasGeometry ?sGeom .
    ?sGeom geo:asWKT ?sWKT .

    ?ev a ev-ont:ElectricVehicleProduct.
    ?ev ev-ont:hasModelType ?model.
    ?ev rdfs:label "Nissan Leaf".
    ?ev ev-ont:hasModelYear "2021"^^xsd:gYear.
    ?ev ev-ont:hasMatchableConnectorType ?co.
    ?co rdfs:label ?co_name.
    VALUES ?co_name{"CHAdeMO" "J1772COMBO" "TESLA"}}}
 \end{lstlisting}

\begin{lstlisting}[
language=SPARQL,
captionpos=b, 
caption={\yanlin{Query for the annual variation trend of the direct-current fast charging (DCFC) infrastructure numbers in New Jersey.}}, 
label={q:q4a},
]
SELECT ?co (SUM(?charger_n) AS ?zip_dcfc_num) ?year
WHERE{
SELECT DISTINCT ?charger_conn ?co ?charger_n ?year WHERE { 
    ?zip a kwg-ont:ZipCodeArea.
    ?state a kwg-ont:AdministrativeRegion_2.
    ?state rdfs:label "New Jersey".
    ?state kwg-ont:sfContains ?zip.

    ?stn a ev-ont:ChargingStation.
    ?stn kwg-ont:sfWithin ?zip.
    ?stn ev-ont:hasOpenYear ?year.
    ?stn ev-ont:hosts ?charger_conn.
    ?charger_conn ev-ont:hasAmount ?charger_n.
    ?charger_conn ev-ont:hasChargerType ?charger.
    VALUES (?charger) {(evr:chargertype.DCFastCharger)}
    ?charger_conn ev-ont:hasConnectorType ?co.

}}  Group By ?co ?year ?stn
 \end{lstlisting}

\begin{lstlisting}[
language=SPARQL,
captionpos=b, 
caption={\yanlin{Query for the annual variation trend of the EV registration numbers in New Jersey. }}, 
label={q:q4b},
]
    SELECT ?co_name ?reg_year (SUM(?evtg_n) AS ?ev_with_dc_num) WHERE{
    SELECT Distinct ?evtg ?evtg_n ?co_name ?reg_year WHERE { 
    ?zip a kwg-ont:ZipCodeArea.
    ?state a kwg-ont:AdministrativeRegion_2.
    ?state rdfs:label "New Jersey".
    ?state kwg-ont:sfContains ?zip.

    ?evtg a ev-ont:ElectricVehicleRegistrationCollection.
    ?evtg ev-ont:hasAmount ?evtg_n.
    ?evtg ev-ont:hasTemporalScope ?reg_year.
    ?evtg ev-ont:hasSpatialScope ?zip.
    ?evtg ev-ont:hasProductInfo ?ev.
    ?ev ev-ont:hasMatchableConnectorType ?co.
    ?co rdfs:label ?co_name.
    VALUES ?co_name{"TESLA" "CHAdeMO" "J1772COMBO"}
}}  GROUP BY ?co_name ?reg_year
 \end{lstlisting}

\begin{lstlisting}[
language=SPARQL,
captionpos=b, 
caption={\yanlin{Query for the annual variation trend of the average charger share per registered EV in New Jersey. Here, we reuse the SPARQL queries from Listing \ref{q:q4a}  and \ref{q:q4b} to constructe a more complex query. }}, 
label={q:q4c},
]
SELECT (?zip_dcfc_num/?ev_with_dc_num AS ?dcfc_per_ev) WHERE{
{``` Query from Listing 4 ```}
{``` Query from Listing 5```}
}
 \end{lstlisting}

\begin{lstlisting}[
language=SPARQL,
captionpos=b, 
caption={\yanlin{Query for the distribution of CCS-EV registrations in New Jersey. }}, 
label={q:q5a},
]
SELECT DISTINCT ?zipcode (SUM(?regNum) AS ?zipRegNum)
WHERE{
        ?zipcode a kwg-ont:ZipCodeArea.
        ?state a kwg-ont:AdministrativeRegion_2.
        ?state rdfs:label "New Jersey".
        ?state kwg-ont:sfContains ?zipcode.
        ?reggroup a ev-ont:ElectricVehicleRegistrationCollection.
        ?reggroup ev-ont:hasSpatialScope ?zipcode.
        ?reggroup ev-ont:hasTemporalScope "2021"^^xsd:gYear.
        ?reggroup ev-ont:hasProductInfo ?ev. 
        ?reggroup ev-ont:hasAmount ?regNum.
        ?ev ev-ont:hasMatchableConnectorType 
            evr:connectortype.J1772COMBO.
	} GROUP BY ?zipcode
}
 \end{lstlisting}

\begin{lstlisting}[
language=SPARQL,
captionpos=b, 
caption={\yanlin{Query for the average CCS-charger share per CCS-EV in New Jersey. We reuse the SPARQL query in Listing \ref{q:q5a} as part of the current query. }}, 
label={q:q5b},
]
SELECT ?zipcode ?zipChargerNum ?zipRegNum 
(?zipChargerNum/?zipRegNum AS ?ratio) 
WHERE{
    ### part 1 the EVSE number at zip code level
    {SELECT DISTINCT ?zipcode (SUM(?chargerNum) AS ?zipChargerNum) 
    WHERE{
    ?zipcode a kwg-ont:ZipCodeArea.
    ?state a kwg-ont:AdministrativeRegion_2.
    ?state rdfs:label "New Jersey".
    ?zipcode kwg-ont:sfWithin ?state.
    ?station a ev-ont:ChargingStation.
    ?station kwg-ont:sfWithin ?zipcode.
    ?station ev-ont:hosts ?chargerCollection. 
    ?chargerCollection ev-ont:hasAmount ?chargerNum.
    ?chargerCollection ev-ont:hasConnectorType 
        evr:connectortype.J1772COMBO.
    } GROUP BY ?zipcode}
    ### part 2 the registration number at zip code level
    {```Query from Listing 7```}
    }
 \end{lstlisting}

\begin{lstlisting}[
language=SPARQL,
captionpos=b, 
caption={\yanlin{Query for distribution of the zip code areas in New Jersey which have average CCS-charger share per CCS-EV less than 0.1. We reuse the SPARQL query in Listing \ref{q:q5b} as part of the current query.}}, 
label={q:q6a},
]
## Condition 1: Average charging resource less than 0.1
SELECT ?zipcode ?transline ?ratio 
WHERE{
    {FILTER(?ratio < 0.1)}
    {```Query from Listing 8```} 
    {SELECT DISTINCT ?zipcode ?transline WHERE {		
        ?zipcode a kwg-ont:ZipCodeArea.
        ?state a kwg-ont:AdministrativeRegion_2.
        ?state rdfs:label "New Jersey".
        ?state kwg-ont:sfContains ?zipcode.
        ?transline a ev-ont:TransmissionLine.
        ?transline kwg-ont:sfCrosses ?zipcode.
        ?transline ev-ont:hasVoltageClass ?v_class.
        ?v_class rdfs:label "500".
    } GROUP BY ?zipcode ?transline}}
 \end{lstlisting}

\begin{lstlisting}[
language=SPARQL,
captionpos=b, 
caption={\yanlin{Query for distribution of the zip code areas in New Jersey which have EV registration number larger than 98. We reuse the SPARQL query in Listing \ref{q:q5a} as part of the current query.}}, 
label={q:q6b}]
## Condition 2: Electric vehicle registration more than 98
SELECT ?zipcode
WHERE{
    {FILTER(?zipRegNum>98)}
    {``` Query from Listing 7```}
    {SELECT DISTINCT ?zipcode ?transline WHERE {		
    ?zipcode a kwg-ont:ZipCodeArea.
    ?state a kwg-ont:AdministrativeRegion_2.
    ?state rdfs:label "New Jersey".
    ?state kwg-ont:sfContains ?zipcode.
    ?transline a ev-ont:TransmissionLine.
    ?transline kwg-ont:sfCrosses ?zipcode.
    ?transline ev-ont:hasVoltageClass ?v_class.
    ?v_class rdfs:label "500".
    } GROUP BY ?zipcode ?transline}}
 \end{lstlisting} 

\end{document}